\documentclass{article}

 \PassOptionsToPackage{numbers, compress}{natbib}

\usepackage[preprint]{neurips_2022}
\usepackage{hyperref}
\usepackage{graphicx}
\usepackage{float}
\usepackage{amsmath, amsthm, amssymb, mathtools, thmtools}

\usepackage[utf8]{inputenc} 
\usepackage[T1]{fontenc}    
\usepackage{hyperref}       
\usepackage{url}            
\usepackage{booktabs}       
\usepackage{amsfonts}       
\usepackage{nicefrac}       
\usepackage{microtype}      
\usepackage{xcolor}         

\title{On the Role of Neural Collapse in Meta Learning Models for Few-shot Learning}

\author{
  Saaketh Medepalli $^{1}$ \quad Naren Doraiswamy $^{2}$ \\ \\
  $^{1}$ Carnegie Mellon University \\ $^{2}$ University of Michigan\\\\
  \texttt{smedepal@cs.cmu.edu}  \quad \texttt{narend@umich.edu} \\
}

\begin{document}

\maketitle

\begin{abstract}
Meta-learning frameworks for few-shot learning aims to learn models that can learn new skills or adapt to new environments rapidly with a few training examples. This has led to the generalizability of the developed model towards new classes with just a few labelled samples. However these networks are seen as black-box models and understanding the representations learnt under different learning scenarios is crucial. Neural collapse ($\mathcal{NC}$) is a recently discovered phenomenon which showcases unique properties at the network proceeds towards zero loss. The input features collapse to their respective class means, the class means form a Simplex equiangular tight frame (ETF) where the class means are maximally distant and linearly separable, and the classifier acts as a simple nearest neighbor classifier. While these phenomena have been observed in simple classification networks, this study is the first to explore and understand the properties of neural collapse in meta learning frameworks for few-shot learning. We perform studies on the Omniglot dataset in the few-shot setting and study the neural collapse phenomenon. We observe that the learnt features indeed have the trend of neural collapse, especially as model size grows, but to do not necessarily showcase the complete collapse as measured by the $\mathcal{NC}$ properties. 
\end{abstract}

\section{Introduction}
Human vision has the innate capability of recognizing new categories when a person is shown just a few samples of that category. For instance, when a person is shown a couple of images of an unseen person or an unseen category, he can recognize the new face quickly by implicitly drawing connections from the acquired prior knowledge.  Although deep neural networks trained on millions of images have in some cases exceeded human performance in large-scale image recognition \cite{he15}, under an open-world setting with emerging new categories it remains a challenging problem how to continuously expand the capability of an intelligent agent from limited new samples, also known as few-shot learning \cite{wang19}.

Moreover, in many machine learning applications, training data and labels are limited and require collection/annotation of new data which can be  prohibitive \cite{hoffman19, wang19}. However, modern machine learning models including deep neural networks often require large amounts of training data to learn good representations needed for downstream tasks. As a result, models that can learn how to solve tasks with little training data are desirable. Meta-learning essentially aims to  solve this issue without having to re-train a base model on the data from the new classes \cite{hospedales20}.

Few-shot classification is an instantiation of meta-learning in the field of supervised learning. After splitting into training and testing sets (with different classes), each dataset $D$ is split into two parts, a support set $S$ for learning and a query/prediction set B for training or testing, $D = \langle S, B \rangle$. Often we consider a $K$-shot $N$-way classification task: the support set contains $K$ labelled examples for each of $N$ classes. Intuitively, the goal of the model is to learn from the small subset of data, i.e: the support set $S$ should classify the data points in the query set $B$ effectively \cite{hospedales20}.

\subsection{Meta-Learning}
\label{meta-learning}
There are three major ways of addressing the learning from limited labeled data in general. They are configured into metric-based, model-based and optimization-based learning methods. The model-based meta learning algorithms design a model specifically for fast learning. They tend to use external/internal memory capabilities of a system to adapt for fast learning \cite{weng2018metalearning}. Memory augmented networks \cite{santoro2016meta} and Meta-networks \cite{DBLP:journals/corr/MunkhdalaiY17} are prime examples for model-based methods. The optimization-based methods aims to adjust the optimization algorithm so that the model can be good at learning with just a few labeled examples. It deals primarily with the modification of the gradient descent optimization algorithm for faster learning \cite{weng2018metalearning}.

In computer vision, the metric based learning algorithms are the most commonly used algorithms where a distance metric is learned between the support set and the query set features to perform the required task at hand \cite{weng2018metalearning}. Primarily, we will be using the prototypical networks proposed by Snell et al. (2017) \cite{snell2017prototypical}, described in further detail in Section \ref{exp_setup}. 

Meta-learning models train over several epochs, each of which consists of several {\it episodes}. An episode consists of the support set $S$, which is the 'training set' for the episode, while the query set $B$ is the 'testing set' for the episode. As mentioned earlier, the way classification is performed over the query set $B$ varies based on the distance metric. Here lies the key essence behind meta-learning. Rather than training over a mini-batch of training data examples, in meta-learning, an epoch comprises a mini-batch of episodes, allowing the model to 'learn how to learn'. 

\subsection{Neural Collapse}

Recent work by Papyan et al. (2020) \cite{papyan} examined the representational properties of deep neural networks and learned that the class features of the final hidden layer associated with training data tend to collapse to the respective class feature means ($\mathcal{NC}_1$). This in turn simplifies the behaviour of the last layer classifier to that of a nearest-class center decision rule ($\mathcal{NC}_4$). The class means further tend to form a simplex equiangular tight frame ($\mathcal{NC}_2$) and the the linear layer classifier weights and the features of the training data can be interchangeable leading to self-duality ($\mathcal{NC}_3$). These properties were termed to be the $\mathcal{NC}$ properties (\cite{papyan}). We intend to check whether these specific properties hold for the meta-learning frameworks. Will these properties generalize to the meta-learning based, few-shot learning scenario? We intend to examine this question in this report.    

\section{Related Work}
\label{related_work}
Empirical and heuristic analysis has been the major reason for the exploration of the potential of deep learning networks. Hence understanding the theoretical underpinnings of such models would help us understand the reason behind the good generalization performance observed in such deep networks. Papyan et al. (\cite{papyan}) studies the representational power of the overparameterized networks at the end phase of training and observed interesting properties in the network behavior. These properties are implicitly understood by the general ML community and have been individually explored. For instance, $\mathcal{NC}_1$ and $\mathcal{NC}_4$ are individually observed in \cite{Giryes2016} and \cite{Cohen2018} respectively. Constraining
the network weights to be tight frames \cite{Cisse} and reducing intra-class variance has also been studied \cite{Mallat}. \\

However, these observations have been studied individually and the 4 of the $\mathcal{NC}$ properties are naturally attained by deep networks without any explicit constraints during training and this functionality makes it a unique feature to analyze and understand. While neural collapse has been evaluated for the transfer learning setting for few-shot learning \cite{galanti2021}, the majorly used meta-learning frameworks have not been analyzed for neural collapse phenomenon.     

\section{Experimental Setup}
\label{exp_setup}

To investigate this phenomenon, we perform several experiments using prototypical networks (ProtoNet) as first proposed by Snell et al. (2017) \cite{snell2017prototypical}. For our datasets, we build on the ProtoNet paper and use the Omniglot dataset \cite{lake15}, specifically developed for few-shot learning. Our code (based on original ProtoNet paper) can be found at \href{https://github.com/saakethmm/NC-prototypical-networks}{\texttt{https://github.com/saakethmm/NC-prototypical-networks}}.

\subsection{Prototypical Networks}

As mentioned earlier, we will be using a form of a metric-based meta-learning model known as the Prototypical Network, first proposed by Snell et al. (2017) \cite{snell2017prototypical}. Within each episode, the mean support set feature vector is calculated for each class:

\begin{equation}
\label{mean}
\mathbf{v}_c=\frac{1}{\left|S_c\right|} \sum_{\left(\mathbf{x}_i, y_i\right) \in S_c} f_\theta\left(\mathbf{x}_i\right)
\end{equation}

where $f_\theta$ is the embedding model, $S_{c}$ is the labeled support set data points and ${\left|S_c\right|}$ is the number of support set data points for the respective class (equal to $K$, i.e., number of shots).  

The query set vectors in $\mathbf{u} \in B$ are embedded by the model and the $\ell_2$ distance is taken between $\mathbf{u}$ and $\mathbf{v}_c$ (from Equation \ref{mean}) for $c \in \mathcal{C}$ where $\mathcal{C} = \{1, 2, ..., N\}$. Formally, the model loss is calculated in \ref{loss}.

\begin{equation}
\label{loss}
\mathcal{L}(\theta) = - \log P_\theta({y = c|\mathbf{u}}) = -\log \frac{\exp{(-\| f_\theta (\mathbf{u}) - \mathbf{v}_c \| _2)}}{\sum_{c' \in \mathcal{C}} \exp{(-\| f_\theta (\mathbf{u}) - \mathbf{v}_c \| _2)}}
\end{equation}

\subsubsection{Backbone Architectures}
Using the loss function in Equation \ref{loss}, two different model archictures were used. The first is a simple convolutional (referred to as ConvNet hereafter) backbone from the ProtoNet paper consisting of four convolutional blocks ($Conv \rightarrow BatchNorm \rightarrow ReLU \rightarrow MaxPool$). 

The other model tested is ResNet-18 for CIFAR-10/100 \cite{he15_resnet}.The number of parameters is shown in Table \ref{num_params}. The implementations of these models are inspired by He et al. (2015) \cite{he15_resnet} but are borrowed from \href{https://github.com/kuangliu/pytorch-cifar}{\texttt{https://github.com/kuangliu/pytorch-cifar}}.

\begin{table}[H]
  \caption{Number of Parameters in Backbones}
  \label{num_params}
  \centering
  \begin{tabular}{lll}
    \toprule
    \cmidrule(r){1-2}
    Name           & \# of Parameters      \\
    \midrule
    ConvNet       & ~111.2k   \\
    ResNet-18     & ~11.17M  \\
    \bottomrule
  \end{tabular}
\end{table}

\subsubsection{Neural Collapse Metrics}
\label{metrics}

We evaluate the neural collapse on two different metrics. The intra-class variability and the formation of simplex equiangular tight frame is estimated by the given two formulae below. It is worth keeping in mind throughout that $K$ is the number of samples or shots while $N$ is the number of classes, perhaps contrary to usual convention.
\\ \\
First, we define the global mean $\overline{\boldsymbol{h}}_G$ and class mean $\overline{\boldsymbol{h}}_n$ of the last layer features ${h_{n,k}}$ as

$$
\overline{\boldsymbol{h}}_G=\frac{1}{K N} \sum_{n=1}^N \sum_{k=1}^K \boldsymbol{h}_{n, k} \quad \overline{\boldsymbol{h}}_n=\frac{1}{K} \sum_{k=1}^K \boldsymbol{h}_{n, k}(1 \leq n \leq N)
$$

Next, we calculate the between class and in-class covariance matrix as

$$
\boldsymbol{\Sigma}_W:=\frac{1}{K N} \sum_{n=1}^N \sum_{k=1}^K\left(\boldsymbol{h}_{n, k}-\overline{\boldsymbol{h}}_n\right)\left(\boldsymbol{h}_{n, k}-\overline{\boldsymbol{h}}_n\right)^{\top}, \quad \boldsymbol{\Sigma}_B:=\frac{1}{N} \sum_{n=1}^N\left(\overline{\boldsymbol{h}}_n-\overline{\boldsymbol{h}}_G\right)\left(\overline{\boldsymbol{h}}_n-\overline{\boldsymbol{h}}_G\right)^{\top}
$$

The variability collapse metric is then measured as the magnitude between-class covariance matrix 
  $\boldsymbol{\Sigma}_B$ and in-class covariance matrix $\boldsymbol{\Sigma}_W$. 

The within class variability collapse can be calculated as (\cite{zhu}):

$$
\mathcal{N} \mathcal{C}_1:=\frac{1}{N} \operatorname{trace}\left(\boldsymbol{\Sigma}_W \boldsymbol{\Sigma}_B^{\dagger}\right)
$$

The between-class variability collapse is calculated as (important to note that ProtoNets have no last-layer classifier, so the last layer-features $\boldsymbol{H} = [\overline{\boldsymbol{h}}_1, \overline{\boldsymbol{h}}_2, ..., \overline{\boldsymbol{h}}_N]$ were used) (\cite{zhu}):

$$
\mathcal{N C}_2:=\left\|\frac{\boldsymbol{H} \boldsymbol{H}^{\top}}{\left\|\boldsymbol{H} \boldsymbol{H}^{\top}\right\|_F}-\frac{1}{\sqrt{N-1}}\left(\boldsymbol{I}_N-\frac{1}{N} \mathbf{1}_N \mathbf{1}_N^{\top}\right)\right\|_F 
$$

\subsection{Datasets}
As mentioned before, the main dataset under consideration is the Omniglot dataset \cite{lake15}. The dataset consists of 1623 distinct, black and white handwritten characters from 50 alphabets. Given the number of input channels in ResNet is 3, the Omniglot images were repeated across the channel dimension (not done for the ConvNet). 

\section{Experimental Results}
\label{results}

The experiments were conducted using ProtoNets on the Omniglot dataset, though using different backbones. We study the training accuracy/loss along with the $\mathcal{NC}_1$/$\mathcal{NC}_2$ metrics on the Omniglot dataset \cite{lake15}. 

For all the experiments, the hyperparameters were left unchanged from the original ProtoNet. During training time, $N_{support} = N_{query} = 60$ and $K_{support} = K_{query} = 5$ per episode. During validation/test-time, $N_{support} = N_{query} = 5$, $K_{support} = 5$, $K_{query} = 15$. Batch size (number of episodes per epoch) is 100. The Adam optimizer was used, with learning rate $\alpha = 0.001$ and decay every 20 epochs. No weight decay regularization was employed. 

\subsection{Few-shot Classification with ConvNet}
\begin{figure}[H]
  \includegraphics[width=\linewidth]{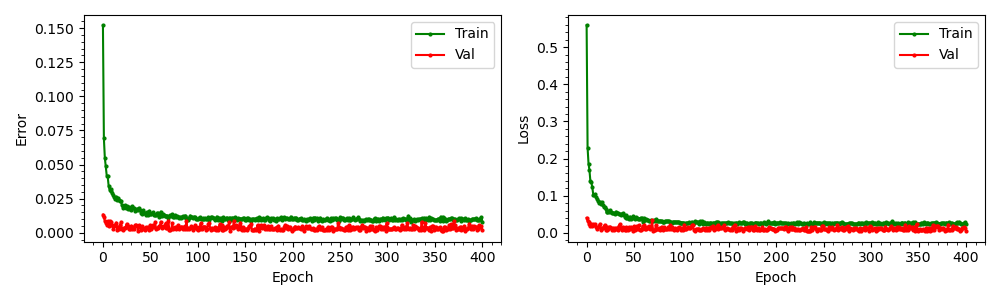}
  \caption{ \small The training and validation error (left) for the ConvNet-based ProtoNet, where error is based on the average number of misclassified query set examples. The training and validation loss (right) show a similar trend. The terminal phase of training is reached around the 100$^{th}$ epoch, though zero training error is never reached.}
  \label{omniglot_loss_acc}
\end{figure}

Prototypical networks with the ConvNet backbone were trained on the Omniglot dataset, as described in \ref{exp_setup}. The results on the training and validation sets can be seen in \ref{omniglot_loss_acc}. It is interesting to note that the validation curves remain below the training curves, suggesting the model is underfitting the training data. 

When testing neural collapse on the metrics defined in Section \ref{metrics}, we note that neither $\mathcal{NC}_1$ nor $\mathcal{NC}_2$ takes place. We make two interesting observations, however. Firstly, both the within-class ($\mathcal{NC}_1$) and between-class variability ($\mathcal{NC}_2$)  drop as training loss decreases to zero, though both stabilize at a trivially non-zero number. It is worth mentioning again that these metrics are measured for {\it each episode} and averaged per epoch, suggesting the model is tending to learn how to form a model feature vector for each class per episode. Secondly, the query set and support set collapse metrics are quite similar, suggesting that the model learns the representation space of the class feature vectors quickly (allowing it to produce a similar structure across both sets of vectors.


\begin{figure}
  \includegraphics[width=\linewidth]{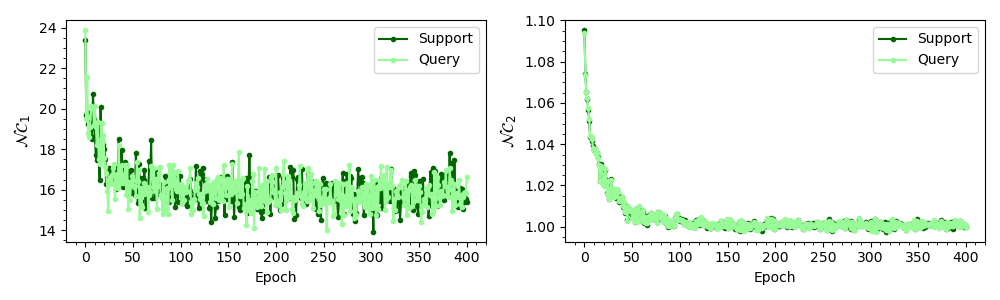}
  \caption{\small The $\mathcal{NC}_1$ score (left) on the support and query sets shows how within-class variability evolves during training for the ConvNet model. The $\mathcal{NC}_2$ score (right) is also shown. Note the noise due to random selection of classes in each episode.} 
  \label{omniglot_nc_train}
\end{figure}

\subsection{Few-shot Classification with ResNet-18}

\begin{figure}
  \includegraphics[width=\linewidth]{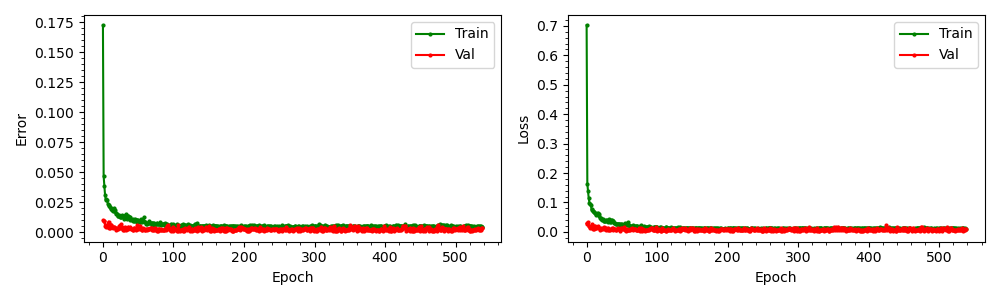}
  \caption{ \small The training and validation error (left) for the ResNet-18-based ProtoNet, where error is based on the average number of misclassified query set examples. The training and validation loss (right) show a similar trend.}
  \label{resnet_loss_acc}
\end{figure}

Given that the ConvNet appears to be underfitting the data, we trained a ProtoNet using ResNet-18, which has \~10 times more parameters (Table \ref{num_params}). As seen in Figure \ref{resnet_loss_acc}, the training error and loss appears to match those of the validation set. 

\begin{figure}
  \includegraphics[width=\linewidth]{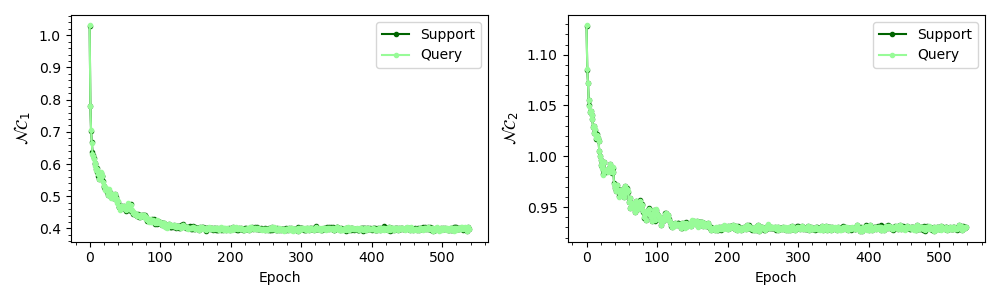}
  \caption{ \small The $\mathcal{NC}_1$ score (left) on the support and query sets shows how within-class variability evolves during training for the ResNet-18 model. The $\mathcal{NC}_2$ score (right) is also shown. Also note the lack of noisiness in the metrics.}
  \label{resnet_nc_train}
\end{figure}

Interestingly, when using a larger model such as ResNet-18, the extent of neural collapse increases (Figure \ref{resnet_nc_train}). Though we still do not observe complete within-class or between-class variability collapse in the embedded features of the support and query sets, the values certainly reach closer to zero. $\mathcal{NC}_1$ also initially begins at a much smaller value compared to Figure \ref{omniglot_nc_train} and reaches \~0.4 before stabilizing, though $\mathcal{NC}_2$ shows only slight decrease in comparison.  

\section{Conclusions}

Neural collapse has been primarily observed in models with the last-layer classifier. Intuitively, such models tune weights to learn strong representations (during training) to achieve similar output scores to the one-hot encoded ground-truth vector. Since metric-based meta-learning methods such as prototypical networks do not use a last-layer classifier and instead use a distance-metric for classification, the phenomenon may not generalize. 

However, from the experiments conducted so far, some version of within-class variability decrease is seen, though this may not qualify as $\mathcal{NC}_1$ as defined by Papyan et al. (2020) \cite{papyan}. Between-class variability decrease is even more difficult to observe. Overall, however, an interesting trend is observed where number of parameters correlates with collapse in prototypical networks. Despite the discrepancy in the method of training, the results observed appear to support the observations made in prior work (\cite{papyan}): overparameterized models lend themselves to stronger neural collapse.

Another interesting observation is that the classification decision in prototypical networks is identical to $\mathcal{NC}_4$, raising the question of whether these models 'force' the model to learn structures with distinct decision boundaries across all combinations of classes. If so, it is interesting to note that the structure learned does not necessarily resemble a Simplex-ETF based on the results, despite the model achieving training loss close to zero.

\section*{Acknowledgements}
We would like to thank Prof. Wei Hu for the opportunity to pursue this work. For guidance on neural
collapse and its metrics, we would also like to thank Prof. Qing Qu and Xiao Li. We also acknowledge the UMich Great Lakes and the UC Berkeley clusters for GPU resources provided for
running experiments.

\newpage

\bibliographystyle{abbrv}
\bibliography{citation}

\appendix

\newpage
\section{Appendix}

\begin{figure}[H]
  \includegraphics[width=\linewidth]{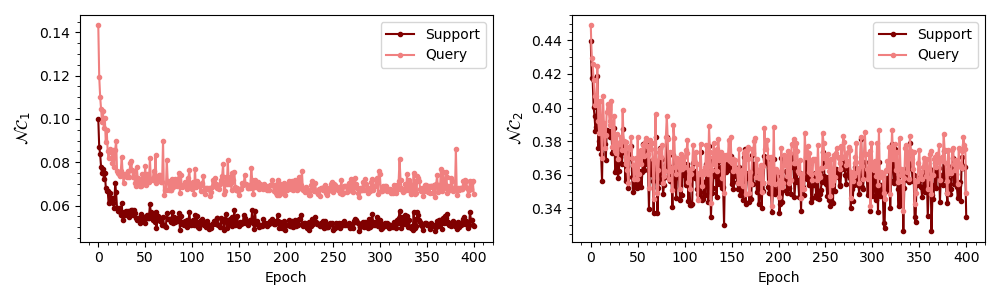}
  \caption{The $\mathcal{NC}_1$ score (left) on the support and query sets shows how within-class variability evolves during validation. The $\mathcal{NC}_2$ score (right) is also shown. Compared to training, the results are far more noisy and smaller in magnitude.} 
  \label{omniglot_nc_val}
\end{figure}

The neural collapse metrics on the validation set for the Omniglot backbone are shown in Figure \ref{omniglot_nc_val}. A similar trend to those seen during training are observed, though the magnitude of values is much smaller. 

Similar observations are made for the ResNet-18 backbone ProtoNet. Building off of the trend from the training-time observations, more collapse seems to be occurring as the model is overparameterized. Moreover, the metrics are less noisy as in the observations made earlier.

\begin{figure}[H]
  \includegraphics[width=\linewidth]{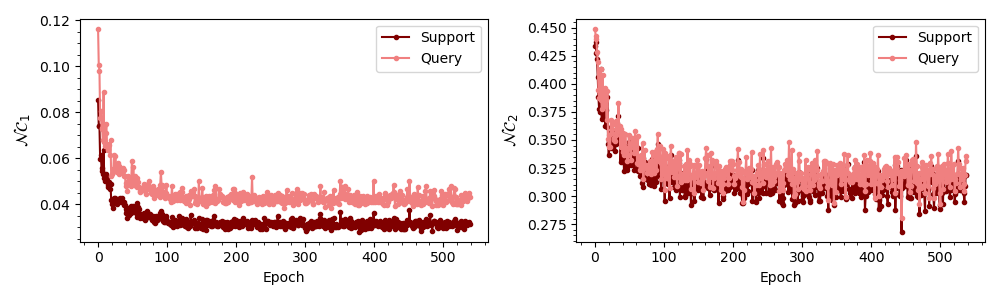}
  \caption{The $\mathcal{NC}_1$ score (left) on the support and query sets shows how within-class variability evolves during validation. The $\mathcal{NC}_2$ score (right) is also shown.} 
  \label{resnet_nc_val}
\end{figure}

\end{document}